# The Convexity and Design of Composite Multiclass Losses

**Mark D. Reid**  MARK.REID@ANU.EDU.AU
**Robert C. Williamson**  BOB.WILLIAMSON@ANU.EDU.AU

Research School of Computer Science, The Australian National University and NICTA, Canberra, Australia

**Peng Sun**  SUNP08@MAILS.TSINGHUA.EDU.CN

Tsinghua National Laboratory for Information Science and Technology(TNList), Department of Automation, Tsinghua University, Beijing 100084, China

## Abstract

We consider composite loss functions for multiclass prediction comprising a proper (*i.e.*, Fisher-consistent) loss over probability distributions and an inverse link function. We establish conditions for their (strong) convexity and explore the implications. We also show how the separation of concerns afforded by using this composite representation allows for the design of families of losses with the same Bayes risk.

## 1. Introduction

We study multiclass *proper composite losses* which are the composition of a *proper loss* and and *invertible link* (both defined formally below). This representation makes the understanding of multiclass losses easier because crucially it seperates two distinct concerns: the statistical and the numerical (Vernet et al., 2011). The statistical properties are controlled by the proper loss. The link function is essentially just a parametrisation. Choice of a suitable link can help — for example, a nonconvex proper loss can be made convex (and thus more amenable to numerical optimisation) by choice of the appropriate link. In this paper we show how this is possible, when a composite loss is convex, and how to convexify an arbitrary proper multiclass loss. The results extend the results on binary composite losses due to Reid & Williamson (2010).

### 1.1. Previous Work

Proper losses are the natural losses to use for probability estimation. The key property of a proper loss (see §2.1 below) is that its expected value is always minimised by the distribution defining the expectation. They have been stud-ied in detail when $n = 2$ (the "binary case") where there is an integral representation (Buja et al., 2005; Gneiting & Raftery, 2007; Reid & Williamson, 2011), and characterization (Reid & Williamson, 2010) when differentiable.

The *theory* of loss functions makes it clear how one ideally chooses a loss — one takes account of one's utility concerning various incorrect predictions (Kiefer, 1987), (Berger, 1985, Section 2.4). The *practice* rarely involves such a step. There is little guidance in the literature concerning how to choose a loss function; typically heuristic arguments are used for the choice — confer *e.g.* (Ighodaro et al., 1982; Nayak & Naik, 1989). Early approaches to multiclass losses used a simple reduction to binary (Dietterich & Bakiri, 1995). More recently, other approaches to the design of losses for multiclass prediction have received attention (Zhang, 2004; Hill & Doucet, 2007; Tewari & Bartlett, 2007; Liu, 2007; Zou et al., 2008; Zhang et al., 2009), although none of these papers developed the connection to proper losses, and most restrict consideration to margin losses (which imply certain symmetry conditions). Zou et al. (2005) proposed a multiclass generalisation of "admissible losses" (their name for classification calibration) for multiclass margin classification. Liu (2007) considered several multiclass generalisations of hinge loss (suitable for multiclass SVMs) and showed some of them were and others not Fisher consistent. When they were not it was shown how the training algorithm could be modified to make the losses behave consistently. Multiclass losses have also been considered in the development of multiclass boosting (see *e.g.* Zhu et al., 2009; Mukherjee & Schapire, 2011; Wu & Lange, 2010).

### 1.2. Key Contribution and Significance

The key point of the paper is as follows. Multiclass losses are necessary in many problems. To date they have typically been constructed as *margin losses* via a convex function $\phi$ applied to a generalised notion of "margin". This is unsatisfactory from a design perspective because it confounds two distinct issues: the decision theoretic notion of a loss (that captures what it is that is important to the end user (Berger, 1985, confer)) and issues associated with the ease of numerical optimisation. Furthermore, margin losses are not particularly well suited to the non-symmetric treatment of different classes, as is necessary in many applications. Fortunately there is a way of neatly separating these two concerns through the use of a *composite loss* (Vernet et al., 2011) where the statistical properties are controlled by the choice of proper loss, and the optimisation properties via the link. This leads to the natural question: suppose one fixes a proper loss (and hence the statistical properties), how should one choose a link to ensure convexity of the overall loss? The answer is not obvious and not trivial, and is the key technical contribution of the paper (Theorem 5). The result opens up the possibility of a more systematic approach to the design of multiclass losses which previously has been approached in a rather ad hoc manner.

## 2. Formal Setup

Suppose $\mathscr{X}$ is some set and $\mathscr{Y} = \{1, \ldots, n\} = [n]$ is a set of labels. We suppose we are given data $S = \{x_i, y_i\}_{i \in [m]}$ such that $y_i \in \mathscr{Y}$ is the label corresponding to $x_i \in \mathscr{X}$. These data follow a joint distribution $\mathbb{P}_{\mathscr{X},\mathscr{Y}}$ on $\mathscr{X} \times [n]$. We denote by $\mathbb{E}_{\mathscr{X},\mathscr{Y}}$ and $\mathbb{E}_{\mathscr{Y}|\mathscr{X}}$ respectively, the expectation and the conditional expectation with respect to $\mathbb{P}_{\mathscr{X},\mathscr{Y}}$. Given a new observation $x$ we want to predict the probability $p_i := \mathbb{P}(Y = i | X = x)$ of $x$ belonging to class $i$, for $i \in [n]$. *Multiclass classification* requires the learner to predict the most likely class of $x$; that is to find $\hat{y} = \arg\max_{i \in [n]} p_i$.

### 2.1. Losses

A loss measures the quality of prediction. Let $\Delta^n := \{(p_1, \ldots, p_n) \colon \sum_{i \in [n]} p_i = 1, \text{and } 0 \leq p_i \leq 1, \forall i \in [n]\}$ denote the *n-simplex*. For multiclass probability estimation, $\ell \colon \Delta^n \to \mathbb{R}_+^n$. For classification, the loss $\ell \colon [n] \to \mathbb{R}_+^n$. The *partial losses* $\ell_i$ are the components of $\ell(q) = (\ell_1(q), \ldots, \ell_n(q))'$ and $\ell_i(q)$ is the loss incurred by predicting $q \in \Delta^n$ when $y = i$. Throughout the paper, $A'$ denotes transpose of a matrix $A$, except when applied to a real-valued function where it denotes derivative. We denote the matrix multiplication of compatible matrices $A$ and $B$ by $A \cdot B$, so the inner product of two vectors $x, y \in \mathbb{R}^n$ is $x' \cdot y$.

The *conditional risk* $L \colon \Delta^n \times \Delta^n \to \mathbb{R}_+$ associated with a loss $\ell$ is the function

$$L(p,q) = \mathbb{E}_{Y \sim p} \ell_Y(q) = p' \cdot \ell(q) = \sum_{i \in [n]} p_i \ell_i(q),$$

where $Y \sim p$ means $Y$ is drawn according to a multinomial distribution with parameter $p \in \Delta^n$. In a typical learning problem one will make an estimate $q \colon \mathscr{X} \to \Delta^n$. The *full risk* is $\mathbb{L}(q) = \mathbb{E}_{\mathscr{X}} \mathbb{E}_{\mathscr{Y}|\mathscr{X}} \ell_Y(q(\mathsf{X}))$.

Minimizing $\mathbb{L}(q)$ over $q \colon \mathscr{X} \to \Delta^n$ is equivalent to minimizing $L(p(x), q(x))$ over $q(x) \in \Delta^n$ for all $x \in \mathscr{X}$ where $p(x) = (p_1(x), \ldots, p_n(x))'$, and $p_i(x) = \mathbb{P}(Y = i | X = x)$. Thus when there is no restriction on the hypothesis class, it suffices to only consider the conditional risk; confer (Reid & Williamson, 2011).

If one is interested in estimating probabilities ($\ell \colon \Delta^n \to \mathbb{R}_+^n$) it is natural to require the associated conditional risk is minimized when estimating the true underlying probability. Such a loss is called *proper* (formally: if $L(p, p) \leq L(p, q)$, $\forall p, q \in \Delta^n$). It is *strictly proper* if the inequality is strict when $p \neq q$ (so it is uniquely minimised by predicting the correct probability). The *conditional Bayes risk*

$$\underline{L} \colon \Delta^n \ni p \mapsto \inf_{q \in \Delta^n} L(p, q).$$

This function is always concave (Gneiting & Raftery, 2007). If $\ell$ is proper, then $\underline{L}(p) = L(p, p) = p' \cdot \ell(p)$. Strictly proper losses induce *Fisher consistent* estimators of probabilities: if $\ell$ is strictly proper, $p = \arg\min_q L(p, q)$.

The losses above are defined on the simplex $\Delta^n$ since the argument (an estimator) represents a probability vector. However it is sometimes desirable to use another set $\mathscr{V}$ of predictions. One can consider losses $\ell \colon \mathscr{V} \to \mathbb{R}_+^n$. Suppose there exists an invertible function $\psi \colon \Delta^n \to \mathscr{V}$. Then $\ell$ can be written as a composition of a loss $\lambda$ defined on the simplex with $\psi^{-1}$. That is, $\ell(v) = \lambda^\psi(v) := \lambda(\psi^{-1}(v))$. Such a function $\lambda^\psi$ is a *composite loss*. If $\lambda$ is proper, we say $\ell$ is a *proper composite loss*, with *associated proper loss* $\lambda$ and *link* $\psi$. Binary proper composite losses have been studied by (Reid & Williamson, 2010).

### 2.2. Matrix Differential Calculus

In order to differentiate the losses we project the $n$-simplex into a subset of $\mathbb{R}^{n-1}$. Let $\tilde{n} := n - 1$. Let

$$\tilde{\Delta}^n := \{(p_1, \ldots, p_{\tilde{n}})' \colon p_i \geq 0, \forall i \in [\tilde{n}], \sum_{i=1}^{\tilde{n}} p_i \leq 1\}$$

denote the *"bottom" of the n-simplex*. We denote by

$$\Pi_\Delta \colon \Delta^n \ni p = (p_1, \ldots, p_n)' \mapsto \tilde{p} = (p_1, \ldots, p_{\tilde{n}})' \in \tilde{\Delta}^n,$$

the projection of the $\Delta^n$, and

$$\Pi_\Delta^{-1} \colon (\tilde{p}_1, \ldots, \tilde{p}_{\tilde{n}})' \mapsto p = (\tilde{p}_1, \ldots, \tilde{p}_{\tilde{n}}, 1 - \sum_{i=1}^{\tilde{n}} \tilde{p}_i)'$$

its inverse.

We use the following notation. The $k$th unit vector $e_k$ is the $n$ vector with all components zero except the $k$th which is 1. The $n$-vector $\mathbb{1}_n := (1, \ldots, 1)'$. The derivative of a function $f$ is denoted $\mathsf{D}f$ and its Hessian $\mathsf{H}f$. The (relative) interior of the simplex is $\mathring{\Delta}^n := \{(p_1, \ldots, p_n) \colon \sum_{i \in [n]} p_i = 1, \text{and } 0 < p_i < 1, \forall i \in [n]\}$ and the boundary is $\partial \Delta^n := \Delta^n \setminus \mathring{\Delta}^n$.

If $A = [a_{ij}]$ is an $n \times m$ matrix, $\text{vec}\, A$ is the vector of columns of $A$ stacked on top of each other. The *Kronecker product* of an $m \times n$ matrix $A$ with a $p \times q$ matrix $B$ is the $mp \times nq$ matrix

$$A \otimes B := \begin{pmatrix} A_{1,1}B & \cdots & A_{1,n}B \\ \vdots & \ddots & \vdots \\ A_{m,1}B & \cdots & A_{m,n}B \end{pmatrix}.$$

We use the following properties of Kronecker products (see Chapter 2 of Magnus & Neudecker (1999)): $(A \otimes B)(C \otimes D) = (AC \otimes BD)$ for all appropriately sized $A, B, C, D$, and $I_1 \otimes A = A$.

If $f \colon \mathbb{R}^n \to \mathbb{R}^m$ is differentiable at $c$ then the *partial derivative* of $f_i$ with respect to the $j$th coordinate at $c$ is denoted $\mathsf{D}_j f_i(c)$ The $m \times n$ matrix of partial derivatives of $f$ is the *Jacobian* of $f$ and denoted

$$(\mathsf{D} f(c))_{i,j} := \mathsf{D}_j f_i(c) \quad \text{for } i \in [m], j \in [n].$$

If $F$ is a matrix valued function $\mathsf{D} F(X) := \mathsf{D} f(\text{vec}\, X)$ where $f(X) = \text{vec}\, F(X)$.

We will require the *product rule* for matrix valued functions (Vetter, 1970; Fackler, 2005): Suppose $f \colon \mathbb{R}^n \to \mathbb{R}^{m \times p}$, $g \colon \mathbb{R}^n \to \mathbb{R}^{p \times q}$ so that $(f \times g) \colon \mathbb{R}^n \to \mathbb{R}^{m \times q}$. Then

$$\mathsf{D}(f \times g)(x) = (g(x)' \otimes I_m) \cdot \mathsf{D} f(x) + (I_q \otimes f(x)) \cdot \mathsf{D} g(x). \tag{1}$$

The *Hessian* at $x \in \mathcal{X} \subseteq \mathbb{R}^n$ of a real-valued function $f \colon \mathcal{X} \to \mathbb{R}$ is the $n \times n$ real, symmetric matrix of second derivatives at $x$

$$(\mathsf{H} f(x))_{j,k} := \mathsf{D}_{k,j} f(x) = \frac{\partial^2 f}{\partial x_k \partial x_j}.$$

Note that the derivative $\mathsf{D}_{k,j}$ is in row $j$, column $k$. It is easy to establish that the Jacobian of the transpose of the Jacobian of $f$ is the Hessian of $f$. That is,

$$\mathsf{H} f(x) = \mathsf{D} \left((\mathsf{D} f(x))'\right) \tag{2}$$

(Magnus & Neudecker, 1999). If $f \colon \mathcal{X} \to \mathbb{R}^m$ for $\mathcal{X} \subseteq \mathbb{R}^n$ is a vector valued function then the Hessian of $f$ at $x \in \mathcal{X}$ is the $mn \times n$ matrix that consists of the Hessians of the functions $f_i$ stacked vertically:

$$\mathsf{H} f(x) := \begin{pmatrix} \mathsf{H} f_1(x) \\ \vdots \\ \mathsf{H} f_m(x) \end{pmatrix}.$$

If $A$ and $B$ are square matrices, $A \succcurlyeq B$ if $A - B$ is positive semidefinite.

## 3. Derivatives of Composite Losses

In order to establish the convexity and other properties of composite losses we start by proving some identities for their first and second derivatives.

Suppose $\ell = \lambda \circ \psi^{-1}$ is composed of the proper loss $\lambda \colon \Delta^n \to \mathbb{R}_+^n$ and the inverse of the link $\psi \colon \Delta^n \to \mathcal{V}$. In order to simplify matters, derivatives for the function $\ell \colon \mathcal{V} \to \mathbb{R}_+^n$ we will assume the set $\mathcal{V}$ is a flat, $(n-1)$-dimensional, convex subset of $\mathbb{R}_+^n$. We do so since if $\mathcal{V}$ were some arbitrary manifold the extra definitions required to make sense of convexity (*e.g.*, in terms of geodesics) and derivatives on manifolds would obscure the thrust of the results below. Furthermore, little is lost either practically or theoretically by assuming a simple $\mathcal{V}$. In practice, predictions are usually vectors in $\mathbb{R}_+^n$, and in theory one could always choose a parametrisation of $\mathcal{V}$ in terms of some simpler space $\mathcal{U}$ and redefine the link via composition with that parametrisation. Alternatively, since links must be invertible, a composite loss could be defined by a choice of loss and choice of *inverse link* $\psi^{-1} \colon \mathcal{V} \to \Delta^n$ for a $\mathcal{V}$ assumed to be flat, etc.

Let $v \in \mathcal{V}$ fixed but arbitrary with corresponding $\tilde{p} = \tilde{\psi}^{-1}(v)$ where $\tilde{\psi}(\tilde{p}) := \psi(\tilde{p}_1, \ldots, \tilde{p}_{\tilde{n}}, p_n)$ with $p_n := \sum_{i=1}^{\tilde{n}} \tilde{p}_i$ is the induced function from $\tilde{\Delta}^n$ to $\mathcal{V}$. By the chain rule and the inverse function theorem the derivatives for each of the partial losses $\ell_i$ satisfy

$$\mathsf{D} \ell_i(v) = \mathsf{D} \left[\lambda_i(\tilde{\psi}^{-1}(v))\right] = \mathsf{D} \lambda_i(\tilde{p}) \cdot [\mathsf{D} \tilde{\psi}(\tilde{p})]^{-1}. \tag{3}$$

Let us write $e_i^n$ as the $i$th $n$-dimensional unit vector, $e_i^n = (0, \ldots, 0, 1, 0, \ldots, 0)'$ when $i \in [n]$, and define $e_i^n = 0_n$ when $i > n$. We can now write $\mathsf{D} \lambda_i(\tilde{p})$ in terms of the $n \times \tilde{n}$ matrix $\mathsf{D} \lambda(\tilde{p})$ using $\mathsf{D} \lambda_i(\tilde{p}) = (e_i^n)' \cdot \mathsf{D} \lambda(\tilde{p})$. Now $\mathsf{D} \lambda(\tilde{p}) = (\mathsf{D} \tilde{\lambda}(\tilde{p})', \mathsf{D} \lambda_n(\tilde{p})')'$, where $\tilde{\lambda}(\tilde{p}) = (\lambda_1(\tilde{p}), \ldots, \lambda_{\tilde{n}}(\tilde{p}))'$, and so

$$\mathsf{D} \lambda_i(\tilde{p}) = (e_i^n)' \cdot \mathsf{D} \lambda(\tilde{p}) = (e_i^n)' \cdot \begin{pmatrix} \mathsf{D} \tilde{\lambda}(\tilde{p}) \\ \mathsf{D} \lambda_n(\tilde{p}) \end{pmatrix}. \tag{4}$$

Furthermore, since $\lambda$ is proper, Lemma 5 by van Erven et al. (2011) means we can use the relationship between a proper loss and its projected Bayes risk $\tilde{\underline{L}} := \underline{L} \circ \Pi_\Delta^{-1}$ to write

$$\mathsf{D} \tilde{\lambda}(\tilde{p}) = W(\tilde{p}) \cdot \mathsf{H} \tilde{\underline{L}}(\tilde{p}) \tag{5}$$

$$\mathsf{D} \lambda_n(\tilde{p}) = y(\tilde{p})' \cdot \mathsf{D} \tilde{\lambda}(\tilde{p}) \tag{6}$$

where $W(\tilde{p}) := I_{\tilde{n}} - \mathbb{1}_{\tilde{n}} \cdot \tilde{p}'$ and where $y(\tilde{p}) := -\tilde{p}/p_n(\tilde{p})$ and $p_n(\tilde{p}) := 1 - \sum_{i \in [\tilde{n}]} p_i$.

Thus, combining (4–6) we have for all $i \in [\tilde{n}]$

$$\begin{aligned} \mathsf{D} \lambda_i(\tilde{p}) &= (e_i^{\tilde{n}})' \cdot W(\tilde{p}) \cdot \mathsf{H} \tilde{\underline{L}}(\tilde{p}), \\ &= ((e_i^{\tilde{n}})' - (e_i^{\tilde{n}})' \cdot \mathbb{1}_{\tilde{n}} \cdot \tilde{p}') \cdot \mathsf{H} \tilde{\underline{L}}(\tilde{p}) \\ &= (e_i^{\tilde{n}} - \tilde{p})' \cdot \mathsf{H} \tilde{\underline{L}}(\tilde{p}), \tag{7} \end{aligned}$$

and

$$\begin{aligned} D\lambda_n(\tilde{p}) &= y(\tilde{p})' \cdot W(\tilde{p}) \cdot H\underline{\tilde{L}}(\tilde{p}) \\ &= \frac{-1}{p_n(\tilde{p})} \tilde{p}' \cdot (I_{\tilde{n}} - \mathbb{1}_{\tilde{n}} \cdot \tilde{p}') \cdot H\underline{\tilde{L}}(\tilde{p}) \\ &= \frac{-1}{p_n(\tilde{p})} (\tilde{p}' - (1-p_n(\tilde{p}))\tilde{p}') \cdot H\underline{\tilde{L}}(\tilde{p}) \\ &= -\tilde{p}' \cdot H\underline{\tilde{L}}(\tilde{p}). \end{aligned} \qquad (8)$$

Finally, noting that by definition $e_n^{\tilde{n}} = 0$, (8) and (7) can be merged and combined with (3) to obtain the following proposition.

**Proposition 1** *For all $i \in [n]$, $\tilde{p} \in \mathring{\tilde{\Delta}}^n$, and $v = \tilde{\psi}(\tilde{p})$,*

$$D\ell_i(v) = -\left(e_i^{\tilde{n}} - \tilde{p}\right)' \cdot \kappa(\tilde{p}) \qquad (9)$$

*where $\kappa(\tilde{p}) := -H\underline{\tilde{L}}(\tilde{p})\left[D\tilde{\psi}(\tilde{p})\right]^{-1}$.*

Using the definition of the Hessian $H\ell_i = D[D\ell_i']$ and the product rule (1) gives

$$\begin{aligned} D\left[D\ell_i(v)'\right] &= D_v[\overbrace{[D\tilde{\psi}(\tilde{p})']^{-1} \cdot H\underline{\tilde{L}}(\tilde{p})'}^{f(\tilde{p})} \cdot \overbrace{(e_i^{\tilde{n}} - \tilde{p})}^{g(\tilde{p})}] \\ &= \left(\left(e_i^{\tilde{n}} - \tilde{p}\right)' \otimes I_{\tilde{n}}\right) \cdot D_v[f(\tilde{p})'] \\ &\quad + (I_1 \otimes f(\tilde{p})) \cdot D\left(e_i^{\tilde{n}} - \tilde{\psi}^{-1}(v)\right) \\ &= \left(\left(e_i^{\tilde{n}} - \tilde{p}\right)' \otimes I_{\tilde{n}}\right) \cdot D_v\left[H\underline{\tilde{L}}(\tilde{p}) \cdot [D\tilde{\psi}(\tilde{p})]^{-1}\right] \\ &\quad - \left([D\tilde{\psi}(\tilde{p})']^{-1} H\underline{\tilde{L}}(\tilde{p})'\right) \cdot [D\tilde{\psi}(\tilde{p})]^{-1} \end{aligned}$$

where $D_v$ is used to indicate that the derivative is with respect to $v$ even when the terms inside the derivative are expressed using $\tilde{p}$. We have now established the following proposition.

**Proposition 2** *For all $i \in [n]$, $\tilde{p} \in \mathring{\tilde{\Delta}}^n$, and $v = \tilde{\psi}(\tilde{p})$,*

$$\begin{aligned} H\ell_i(v) &= -\left(\left(e_i^{\tilde{n}} - \tilde{p}\right)' \otimes I_{\tilde{n}}\right) \cdot D\left[\kappa\left(\tilde{\psi}^{-1}(v)\right)\right] \\ &\quad + (\kappa(\tilde{p})') \cdot [D\tilde{\psi}(\tilde{p})]^{-1}. \end{aligned}$$

*where $\kappa(\tilde{p}) := -H\underline{\tilde{L}}(\tilde{p}) \cdot [D\tilde{\psi}(\tilde{p})]^{-1}$.*

The product $\kappa(\tilde{p}) := -H\underline{\tilde{L}}(\tilde{p})[D\tilde{\psi}(\tilde{p})]^{-1}$ that appears in both propositions above can be interpreted as the curvature of the Bayes risk function $\underline{\tilde{L}}$ relative to the rate of change of the link function $\tilde{\psi}$. When the link function is the identity function $\tilde{\psi}(\tilde{p}) = \tilde{p}$ (i.e. when we are in the non-composite proper loss case) the expressions for the derivative and Hessian of each $\ell_i$ simplify to

$$D\ell_i(\tilde{p}) = (e_i^{\tilde{n}} - \tilde{p})' \cdot H\underline{\tilde{L}}(\tilde{p}) \qquad (10)$$

$$H\ell_i(\tilde{p}) = \left(\left(e_i^{\tilde{n}} - \tilde{p}\right)' \otimes I_{\tilde{n}}\right) \cdot D\left[H\underline{\tilde{L}}(\tilde{p})\right] - H\underline{\tilde{L}}(\tilde{p})' \qquad (11)$$

The form of $\kappa$ as the product of $H\underline{\tilde{L}}$ and $D\tilde{\psi}$ suggests another simplification. The *canonical link function* for a loss $\lambda$ with Bayes risk $\underline{L}$ is defined by the relationship

$$\tilde{\psi}_\lambda(\tilde{p}) = -D\underline{\tilde{L}}(\tilde{p})'$$

for all $\tilde{p}$. (We will show in section 5.1 that this is guaranteed to be a legitimate link.) We see the term $\kappa$ simplifies to $\kappa(\tilde{p}) = I_{\tilde{n}}$ since $D\tilde{\psi}(\tilde{p}) = -D(D\underline{\tilde{L}}(\tilde{p})') = -H\underline{\tilde{L}}(\tilde{p})$. For this choice of link function, the first and second derivatives become considerably simpler.

**Proposition 3** *If $\lambda : \Delta^n \to \mathbb{R}_+^n$ is a proper loss and $\tilde{\psi}_\lambda$ is its associated canonical link then, for all $i \in [n]$, $\tilde{p} \in \mathring{\tilde{\Delta}}^n$, and $v = \tilde{\psi}_\lambda(\tilde{p})$, the composite loss $\ell = \lambda \circ \tilde{\psi}$ satisfies*

$$D\ell_i(v) = (e_i^{\tilde{n}} - \tilde{p}) \qquad (12)$$

$$H\ell_i(v) = \left[H\underline{\tilde{L}}(\tilde{p})\right]^{-1}. \qquad (13)$$

The simplified form of the Hessian above is established by noting that since $\kappa(\tilde{p}) = I_{\tilde{n}}$ we have $D[\kappa(\tilde{\psi}^{-1}(v))] = 0$ for all $v \in \mathcal{V}$ in Proposition 2.

While the above propositions hold for any number of classes $n$, it is instructive (both here and later in the paper) to examine the binary case where $n = 2$. In this case, Proposition 1 and Proposition 2 reduce to

$$\ell_1'(v) = -(1-p)\kappa(\tilde{p}) \quad ; \quad \ell_2'(v) = p\kappa(\tilde{p}) \qquad (14)$$

$$\ell_1''(v) = \frac{-(1-p)\kappa'(p) + \kappa(p)}{\tilde{\psi}'(p)} \qquad (15)$$

$$\ell_2''(v) = \frac{p\kappa'(p) + \kappa(p)}{\tilde{\psi}'(p)} \qquad (16)$$

where $\kappa(\tilde{p}) = -\frac{\underline{\tilde{L}}''(p)}{\tilde{\psi}'(\tilde{p})} \geq 0$ and so $\frac{d}{dv}\kappa(\tilde{\psi}^{-1}(v)) = \frac{\kappa'(p)}{\tilde{\psi}'(p)}$.

## 4. Convexity

Convexity of a loss is desirable for the ease of numerical optimisation of an empirical risk. We will now consider when multiclass proper losses are convex, and give a characterisation in terms of the corresponding Bayes risk which as we have seen is the natural way to parametrise a loss. The results in this section are the multiclass generalisation of the characterisation of convexity of binary proper losses (Reid & Williamson, 2010). In fact we obtain more general results even in the binary case because here we consider *strongly* convex losses. We will also show how any nonconvex proper loss can be made convex by suitable choice of a link function, specifically: the canonical link.

We define a loss $\ell \colon \Delta^n \to \mathbb{R}_+^n$ to be *convex* if for all $p \in \Delta^n$, the map $\Delta^n \ni q \mapsto p' \cdot \ell(q)$ is convex for all $q$. That is, a loss is convex if, under and distribution $p$ over outcomes $i \in [n]$, the expected loss $\mathbb{E}_{i \sim p}[\ell_i(q)]$ is convex. It is easy to see that $\ell$ is convex if and only if $\ell_i \colon \Delta^n \to \mathbb{R}_+$ is convex for all $i \in [n]$. (The "if" part follows since a sum of convex functions is convex; the "only if" follows by considering $p = e_i$, for $i \in [n]$.)

**Definition 4** *Suppose $C \subset \mathbb{R}^n$ is convex. A function $f \colon C \to \mathbb{R}$ is* strongly convex *on $C$ with modulus $c \geq 0$ if for all $x, x_0 \in C$, $\forall \alpha \in (0, 1)$,*

$$f(\alpha x + (1-\alpha) x_0) \leq \alpha f(x) + (1-\alpha) f(x_0) \\ - \frac{1}{2} c \alpha (1-\alpha) \|x - x_0\|^2.$$

When $c = 0$ in the above definition, $f$ is convex. The function $f$ is strongly convex on $C$ with modulus $c$ if and only if $x \mapsto f(x) - \frac{c}{2}\|x\|^2$ is convex on $C$ (Hiriart-Urruty & Lemaréchal, 2001, page 73). Therefore, the maps $v \mapsto \ell_i(v)$ are $c$-strongly convex if and only if $\mathsf{H}\ell_i(v) \succcurlyeq c I_{\tilde{n}}$. By applying Proposition 2 we obtain the following characterisation of the $c$-strong convexity of the loss $\ell$.

**Theorem 5** *A proper composite loss $\ell = \lambda \circ \psi^{-1}$ is strongly convex with modulus $c \in [0, 1]$ if and only if for all $\tilde{p} \in \mathring{\Delta}^n$ and for all $i \in [n]$*

$$\left( (e_i^{\tilde{n}} - \tilde{p}) \otimes I_{\tilde{n}} \right) \cdot \mathsf{D}\left( \kappa\left( \tilde{\psi}^{-1}(v) \right) \right) \preccurlyeq \kappa(\tilde{p})' \cdot [\mathsf{D}\tilde{\psi}(\tilde{p})]^{-1} - c I_{\tilde{n}}. \tag{17}$$

We now consider the implications of Theorem 5 in two special cases: in the multiclass case with canonical link, and in the binary case with the identity link.

### 4.1. Implications for Canonical Links

Recall that the canonical link $\tilde{\psi}_\ell$ is chosen so that $\tilde{\psi}(\tilde{p}) = -\mathsf{D}\tilde{L}(\tilde{p})'$. This simplifies $\kappa(\tilde{p})$ to the identity matrix $I_{\tilde{n}}$ so $\mathsf{D}\kappa(\tilde{p}) = 0$. In this case, equation (17) reduces to the following corollary.

**Corollary 6** *If $\ell = \lambda \circ \psi^{-1}$ is defined so that $\tilde{\psi} = -\mathsf{D}\tilde{L}'$ then each map $v \mapsto \ell_i(v)$ is $c$-strongly convex if and only if $\left[ -\mathsf{H}\tilde{L}(\tilde{p}) \right]^{-1} \succcurlyeq c I_{\tilde{n}}$, or equivalently $-\mathsf{H}\tilde{L}(\tilde{p}) \preccurlyeq \frac{1}{c} I_{\tilde{n}}$.*

An immediate consequence of this result is obtained by observing the definiteness constraint is always met when $c = 0$ since $\tilde{L}$ is always a concave function. Thus, *using a canonical link guarantees a composite loss is convex.*

There is a analogous upper definiteness condition to strong convexity that has implications for optimisation rates. In (Boyd & Vandenberghe, 2004, §9.1.2) it is shown that if a twice differentiable function $f \colon X \to \mathbb{R}$ satisfies

$$M I \succcurlyeq \mathsf{H}f(x) \succcurlyeq m I$$

for all $x \in X \subset \mathbb{R}^n$ then the value $\frac{M}{m}$ is an upper bound on the *condition number* of $\mathsf{H}f$, that is, the ratio of maximum to minimum eigenvalue of $\mathsf{H}f$. This value measures the eccentricity of the sublevel sets of $f$ and controls the rate at which optima of $f$ are approached.

Applying this result to the Hessian of a composite loss $\ell$ with a canonical link shows that the condition number bound is controlled by the Hessian of the Bayes risk of $\ell$. Specifically, if the condition number is to be no more than $M/m$ then $\frac{1}{M} \succcurlyeq -\mathsf{H}\tilde{L}(\tilde{p}) \succcurlyeq \frac{1}{m}$ for all $\tilde{p}$. In the case that $M = m$ and the condition number is 1, the only Hessian that suffices is $\mathsf{H}\tilde{L}(\tilde{p}) = -I_{\tilde{n}}$ which is easily shown to be the Bayes risk surface for square loss. Thus, *square loss is the only canonical composite loss for which a condition number of 1 is possible.*

### 4.2. Implications for Binary Losses

In the binary case, when $n = 2$, (15) and (16) and the positivity of $\tilde{\psi}'$ simplify (17) to two conditions:

$$\left. \begin{array}{rcl} (1-p)\kappa'(p) & \leq & \kappa(p) - c\tilde{\psi}'(p) \\ -p\kappa'(p) & \leq & \kappa(p) - c\tilde{\psi}'(p) \end{array} \right\}, \ \forall p \in (0,1).$$

Further assuming that $\tilde{\psi}$ is the identity link ($\tilde{\psi}(v) = v$) and letting $w(p) := -\tilde{L}(p)$ gives

$$\left. \begin{array}{rcl} w'(p) & \leq & \frac{1}{1-p}(w(p) - c)) \\ w'(p) & \geq & \frac{-1}{p}(w(p) - c) \end{array} \right\}, \ \forall p \in (0,1)$$

$$\Leftrightarrow -\frac{1}{p} \leq \frac{w'(p)}{w(p) - c} \leq \frac{1}{1-p}, \ \forall p \in (0,1). \tag{18}$$

The last equivalence is achieved by dividing through by $w(p) - c$ which must necessarily be positive since if it were not the final pair of inequalities would imply $-\frac{1}{p} \geq \frac{1}{1-p}$, a contradiction given that $p \in [0,1]$. Note that (18) reduces to (Reid & Williamson, 2010, Corollary 26) for $c = 0$.

Observe that if $g(p) := \log(w(p) - c)$ then $g'(p) = \frac{w'(p)}{w(p)-c}$ is the middle term in (18). This allows a simplification of the inequality. Specifically, if we assume $w(\frac{1}{2}) = 1$ then

$$-\frac{1}{p} \leq g'(p) \leq \frac{1}{1-p}, \ \forall p \in (0,1)$$

$$\Rightarrow \int_{\frac{1}{2}}^{q} -\frac{1}{p} dp \lesseqgtr \int_{\frac{1}{2}}^{q} g'(p) dp \lesseqgtr \int_{\frac{1}{2}}^{q} \frac{1}{1-p} dp, \ \forall q \in (0,1)$$

$$\Leftrightarrow -\log(q) - \log(2) \lesseqgtr g(q) - \log(1-c) \tag{19}$$

$$\lesseqgtr -\log(2) - \log(1-q), \ \forall q \in (0,1)$$

$$\Leftrightarrow \frac{1}{2q} \lesseqgtr e^{g(q) - \log(1-c)} \lesseqgtr \frac{1}{2(1-q)}, \ \forall q \in (0,1)$$

which gives the following proposition purely in terms of $w(p)$, rather than $w(p)$ and its derivative.

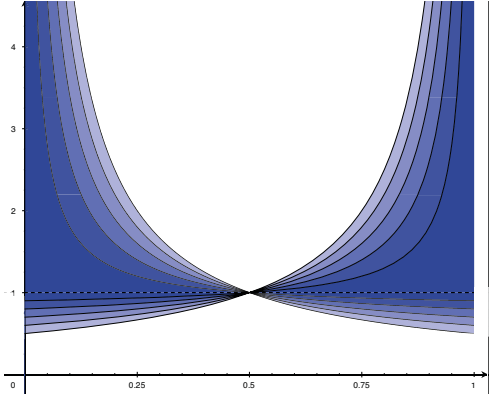

Figure 1. Illustration of range of $w(\tilde{p}) = -\underline{\tilde{L}}''(\tilde{p})$ necessary for a binary proper loss is strongly convex with modulus $c \in \{0, \frac{1}{5}, \frac{2}{5}, \frac{3}{5}, \frac{4}{5}, 1\}$. The regions $R_c$ are nested by subsethood so that $R_0 \supset R_{1/5} \supset R_{2/5} \supset R_{3/5} \supset R_{4/5} \supset R_1$ which is simply the dotted line (containing only the function $w(c) = 1$, $\forall c \in [0,1]$). The palest shaded region corresponds to $R_0$, the allowable range of $w(c)$ necessary for the corresponding proper loss to be convex, and the darkest corresponds to $R_{4/5}$.

**Proposition 7** *Let $w(p) = -\mathsf{H}\underline{\tilde{L}}(p) = -\underline{\tilde{L}}''(p)$ and assume $w(1/2) = 1$. A proper binary loss $\ell\colon \Delta^2 \to \mathbb{R}_+^2$ is strongly convex with modulus $c \in [0,1]$ only if*

$$\frac{1}{2p} \lesseqgtr \frac{w(p) - c}{1 - c} \lesseqgtr \frac{1}{2(1-p)}, \quad \forall p \in (0,1), \qquad (20)$$

*where $\lesseqgtr$ denotes $\leq$ for $p \geq \frac{1}{2}$ and denotes $\geq$ for $p \leq \frac{1}{2}$.*

When $c = 0$ (corresponding to $\ell$ being convex) this is equivalent to an expression by Reid & Williamson (2010, Equation 31). Equation 20 is illustrated in Figure 1.

The above proposition only gives a necessary condition for strong convexity. (In addition to $w$ belonging to the specified region, $w'(p)$ also needs to be suitably controlled). A sufficient condition is useful for designing strongly convex proper losses. Observe that if

$$w(p) = \exp\left(\int_{1/2}^p u(t) dt + K\right) + C$$

where $u\colon [0,1] \to \mathbb{R}$ and $K, c \in \mathbb{R}$, then $\frac{\partial}{\partial p} \log(w(p) - c) = u(p)$. We require $w(1/2) = 1$ thus $\exp\left(\int_{1/2}^{1/2} u(t) dt + K\right) + c = 1$, so $e^K = 1 - c$ and

$$w(p) = (1-c)\exp\left(\int_{1/2}^p u(t) dt\right) + c \qquad (21)$$

satisfies (18) if

$$-\frac{1}{p} \leq u(p) \leq \frac{1}{1-p}, \quad \forall p \in (0,1), \qquad (22)$$

and hence the loss with weight function $w$ is strongly convex with modulus $c$. Thus by choosing $u$ and constructing $w$ via (21) one can design strongly convex proper binary losses.

One can ask whether equation (17) can be simplified in the $n > 2$ case by using a matrix version of the logarithmic derivative trick. Such a result does exist (Horn & Johnson, 1991, Section 6.6.19) but it requires that $(\mathsf{H}\underline{\tilde{L}}(\tilde{p}))^{-1}$ and $\mathsf{D}(\mathsf{H}\underline{\tilde{L}}(\tilde{p}))$ commute for all $\tilde{p} \in \tilde{\Delta}^n$, which is not generally the case.

## 5. Designing Losses

The theory developed above suggests that each choice of proper loss $\lambda$ and link function $\psi$ results in an overall loss function with properties (*e.g.*, convexity) that depend entirely on their relationship to each other. Given these two "knobs" for parameterising a loss function, we can begin to ask what kind of practical trade-offs are involved when selecting a composite loss as a surrogate loss for a particular problem.

We now propose a simple scheme for constructing families of losses with the same Bayes risk. This is achieved by fixing a choice of proper loss $\lambda$ and creating a parameterised family (described below) of link functions $\psi_\alpha$ for parameters $\alpha \in A$. Since the Bayes risk is entirely determined by $\lambda$ any composite loss $\lambda \circ \psi_\alpha^{-1}$ for $\alpha \in A$ will have Bayes risk $\underline{L}(p) = p'\lambda(p)$. Thus, we are able to examine the effect different choices of composite loss can have on a problem *without changing the essential underlying problem*.[1] Through some simple experiments we validate that, at least in the context of boosting, the choice of link can have a significant affect on the convergence and robustness of learning.

### 5.1. Parameterised Links

In order to construct a parametric family of links we first choose some set of inverse link functions $\mathscr{B} = \{\psi_1^{-1}, \ldots, \psi_B^{-1}\}$ with a common domain, that is, $\psi_b^{-1}\colon \mathscr{V} \to \Delta^n$ for a common $n$ and $\mathscr{V}$. This collection will be called the *basis set* of link functions. We then take the convex hull of $\mathscr{B}$ to form a set of inverse link functions $\Psi^{-1} = \mathrm{conv}(\mathscr{B})$. Each $\psi^{-1} \in \Psi^{-1}$ is then identified with the unique $\alpha \in A = \Delta^B$ such that $\sum_{b=1}^B \alpha_b \psi_b^{-1} = \psi^{-1}$. For this construction to be valid, it it necessary to show that every such $\psi^{-1} \in \Psi^{-1}$ is indeed an inverse link function, that is, it is invertible.

The following proposition shows that it suffices to assume that all of the basis functions are *strictly monotonic*. A function $f\colon V \to \mathbb{R}^n$ is *monotone* if for all distinct $u,v \in V$ $(f(u)-f(v))'(u-v) \geq 0$. Strict monotonicity holds when the inequality is strict.

---

[1] Of course, this argument only holds in a point-wise analysis. That is, where choices for estimates $p(x)$ can be made independently. Once a restricted hypothesis class for the functions $p$ is introduced the choice of link can affect the minimal achievable risk. Understanding this interaction is left as future work.

**Proposition 8** *Every function $\psi^{-1}$ in the set $\Psi^{-1} = \mathrm{conv}(\mathscr{B})$ is invertible whenever each basis function in $\mathscr{B}$ is strictly monotone.*

This result is a consequence of: 1) strict monotonicity being preserved under convex combination; and 2) strict monotonicity implies invertibility. The first claim is established by considering strictly monotonic $f$ and $g$ and some $\alpha \in [0,1]$ and noting that if $h = \alpha f + (1-\alpha)g$ then $(h(u)-h(v))'(u-v) = \alpha(f(u)-f(v))'(u-v) + (1-\alpha)(g(u)-g(v))'(u-v) > 0$. A strictly monotone function $f$ that is not invertible is impossible since if we have $(f(u)-f(v))'(u-v) > 0$ for all $u,v$ then a $u \neq v$ s.t. $f(u) = f(v)$ would lead to a contradiction.

Strictly monotone basis functions are easily obtained via canonical links for strictly proper losses. By definition, a canonical link satisfies $\tilde{\psi} = -\mathrm{D}\underline{\tilde{L}}$ for some Bayes risk function. Strict properness guarantees $\underline{\tilde{L}}$ is strictly concave (Vernet et al., 2011) and Kachurovskii's theorem (Showalter, 1997) states that the derivative of a function is (strictly) monotonic if and only if the function is (strictly) convex. Since $(f(f^{-1}(u)) - f(f^{-1}(v)))'(f^{-1}(u) - f^{-1}(v)) = (u-v)'(f^{-1}(u) - f^{-1}(v))$ we see that strictly monotone functions have strictly monotone inverses and we have established the following proposition.

**Proposition 9** *If $\lambda$ is a strictly proper loss then its canonical link $\tilde{\psi}_\lambda = -\mathrm{D}\underline{\tilde{L}}$ has a strictly monotone inverse.*

This result means that a set of basis links can be defined via a choice of strictly concave Bayes risk functions. As an example, the class of Fisher-consistent margin losses proposed by Zou et al. (2008) provides a flexible starting point for designing sets of link functions as described above. They give explicit formulae for the inverse link for a composite loss defined by a choice of convex function $\phi\colon \mathbb{R} \to \mathbb{R}$. Specifically, if the loss for predicting $v \in \mathscr{V} = \{v \in \mathbb{R}^n : \sum_i v_i = 0\}$ is given by $\ell(v) = \phi(v_j)$ then its inverse link is $\psi_\phi^{-1}(v) = \frac{1}{Z_\phi(v)}\left([\phi'(v_i)]^{-1}\right)_{i=1}^n$ where $Z_\phi(v)$ normalises the vector to lie in $\Delta^n$. Each choice of strictly convex $\phi$ gives a valid inverse link which can be used as a basis function.

### 5.2. Experiments

In order to test the impact the choice of link has on the convergence rate we ran a simple experiment using a basic multiclass boosting algorithm, much like the $L_K$-TreeBoost method (Friedman, 2001) for trees with two terminal nodes. In this experiment $\lambda$ was fixed to be the log loss (*i.e.*, $\lambda_i(p) = -\log p_i$), and two basis links $\psi_{\exp}^{-1}$ and $\psi_{\mathrm{sq}}^{-1}$ correspond to choosing in the preceeding subsection the convex functions $\phi(t) = e^{-t}$ and $\phi(t) = (1-t)^2$, respectively. Inverse link functions $\psi_\alpha^{-1} = \alpha \psi_{\exp}^{-1} + (1-\alpha)\psi_{\mathrm{sq}}^{-1}$ were chosen for $\alpha \in \{0, 0.25, 0.5, 0.75, 1\}$ to construct composite losses $\ell_\alpha = \lambda \circ \psi_\alpha^{-1}$. For each loss, boosting was performed on data generated by i.i.d. sampling from three 2-dimensional Gaussians at (0,0), (2,2), and (-2,2) with identity covariance. 4,800 training and 1,200 test samples were used with equal class proportions in both sets. The results shown in Figure 2 clearly indicate the importance of careful link selection.

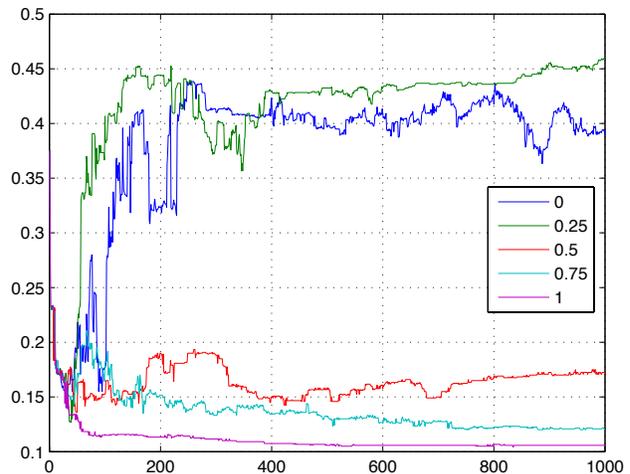

*Figure 2.* Test set accuracy after $T$ iterations of boosting using $\lambda_\alpha$ for $\alpha \in \{0, 0.25, 0.5, 0.75, 1\}$ described in the text.

## 6. Conclusion

Composite multiclass losses are a natural family of losses for multiclass probability estimation and classification which provide a seperation of concerns between the statisitical performance ($\lambda$) and the parametrisation ($\psi$). We have shown that the requirement that the loss decompose into a proper loss and an inverse link gives enough structure to obtain simple expressions for the gradient and derivative of these losses, especially in the binary case or under the additional assumption that the link be *canonical* for the loss. We used these results to provide sufficient conditions for the convexity and condition numbers for composite losses and a general scheme for designing families of multiclass losses with the same Bayes risk. Preliminary experiments show that there are trade-offs inherent when designing multiclass losses that require further investigation.

## Acknowledgements

This work was supported by the Australian Research Council (ARC). NICTA is funded by the Australian Government